%% file: main.tex
\documentclass{article}

\usepackage[main,final]{neurips_2026}

\usepackage[utf8]{inputenc}
\usepackage[T1]{fontenc}
\usepackage{hyperref}
\usepackage{url}
\usepackage{booktabs}
\usepackage{amsfonts}
\usepackage{amsmath,amssymb,amsthm,mathtools}
\usepackage{nicefrac}
\usepackage{microtype}
\usepackage{xcolor}
\usepackage{enumitem}
\usepackage{color-edits}
\addauthor[Mahdi]{m}{blue}

\theoremstyle{plain}
\newtheorem{theorem}{Theorem}[section]
\newtheorem{lemma}[theorem]{Lemma}
\newtheorem{corollary}[theorem]{Corollary}

\theoremstyle{definition}
\newtheorem{definition}[theorem]{Definition}
\newtheorem{assumption}[theorem]{Assumption}

\theoremstyle{remark}
\newtheorem{remark}[theorem]{Remark}

\newcommand{\Pa}{\mathrm{Pa}}

\newcommand{\Span}{\mathrm{span}}
\newcommand{\argmin}{\mathop{\mathrm{arg\,min}}}
\newcommand{\norm}[1]{\left\lVert #1\right\rVert}
\newcommand{\ip}[2]{\left\langle #1,#2\right\rangle}
\newcommand{\Risk}{\mathcal R}
\newcommand{\EmpRisk}{\widehat{\mathcal R}}

\newcommand{\SkRisk}{\widetilde{\mathcal R}}
\newcommand{\Hlin}{\mathcal H}
\newcommand{\epsCtwo}{\varepsilon_{\mathrm{C2}}}

\title{Finite-Sample and Communication-Efficient Networked Information Aggregation}

\author{%
  Mohammadhossein Bateni \\
  Google Research \\
  \texttt{bateni@google.com}
  \And
  Zahra Hadizadeh \\
  University of California, \\
  Irvine \\
  \texttt{zhadizad@uci.edu}
  \And
  MohammadTaghi Hajiaghayi \\
  University of Maryland \\
  \texttt{hajiagha@umd.edu}
  \AND
  Mahdi JafariRaviz \\
  University of Maryland \\
  \texttt{mahdij@umd.edu}
  \And
  Shayan Taherijam \\
  University of California, \\
  Irvine \\
  \texttt{staherij@uci.edu}
}

\begin{document}

\maketitle

\begin{abstract}
Building on SODA 2026 work \citet{kearns2025networked}, we study learning in directed
acyclic graphs in which each agent observes only a subset of the input
coordinates and may use the predictions of its parents as additional
features. In the finite-sample protocol, the message sent along each
edge is the agent's prediction vector on a shared sample, so the
communication cost is linear in the sample size \(m\). In this paper we
study this model under communication constraints.
First, we revisit the finite-sample analysis of the full-message
protocol. We identify a gap in the finite-sample proof of
\citet{kearns2025networked} and develop a different argument that
establishes a finite-sample population-risk guarantee along paths
satisfying deterministic \(M\)-coverage.
Second, we introduce a shared-sketch protocol in which each agent sends
a \(k\)-dimensional linear sketch of its prediction vector instead of
the full \(m\)-dimensional message, where \(k\ll m\). We analyze this
protocol through a reduction to least squares on the compressed sample
together with an oblivious subspace embedding argument. The resulting
bound separates statistics from communication: the statistical term is
still governed by the original sample size \(m\), while the additional
error due to compression is controlled by the sketch dimension \(k\).
For Gaussian sketches, we obtain an explicit high-probability
population-risk bound, showing that row subsampling is not the only
possible communication-statistics tradeoff for networked information
aggregation.
\end{abstract}

\input{introduction}
\input{preliminaries}
\input{finite_sample}
\input{sketching}

\begin{ack}
The work is partially supported by DARPA expMath, ONR MURI 2024 award on Algorithms,
Learning, and Game Theory, Army-Research Laboratory (ARL) grant W911NF2410052, NSF
AF:Small grants 2218678, 2114269, 2347322. The authors declare no competing interests.
\end{ack}

\bibliography{references}
\bibliographystyle{plainnat}

\input{appendix}

\end{document}

%% file: introduction.tex
\section{Introduction}
\label{sec:introduction}

In many distributed learning problems, no single agent observes all of
the input features. Instead, each agent sees only a subset of the
coordinates and can communicate only with nearby agents in a network. A
basic question is whether such local communication is enough to recover
the performance of a learner that has access to all features at once.

\citet{kearns2025networked} study this question for linear regression on
directed acyclic graphs. In their model, agent \(i\) observes a subset
\(S_i\subseteq[d]\) of the coordinates and may use the predictions sent
by its parents. Their main analysis is population-based: each agent
solves a least-squares problem with respect to the underlying
distribution, and along a path they show that if every \(M\) consecutive
agents together observe all \(d\) features, then the network can compete
with the best predictor that uses all coordinates.

Our focus is the communication needed to retain this aggregation benefit
on a finite shared sample. The model is motivated by vertically
partitioned learning on aligned examples and stacked prediction systems
that pass upstream fitted values as downstream features.

It is important to be clear about what is communicated in this model.
The message passed by an agent is not just a coefficient vector. Rather,
the message is the agent's prediction itself, which later agents may use
as an additional feature in their own least-squares problems. In a
finite table, this corresponds to passing the full column of fitted
values. In the population analysis, the analogous message is the
prediction as an object defined over the data distribution rather than a
fixed finite-dimensional vector. This is one reason why communication is
not directly quantified in the population model.

Appendix~C of \citet{kearns2025networked} considers a finite-sample
version of this model. A sample
\[
    D_m=\{(x^{(1)},y^{(1)}),\dots,(x^{(m)},y^{(m)})\}
\]
is shared across the network, each agent still sees only its own feature
subset \(S_i\), and the least-squares problem at each node is solved on
these sampled rows. In this setting, the message sent by an agent is its
prediction vector on the \(m\) sampled rows. Thus the recursive learning
rule is the same in spirit, but communication is now represented by an
explicit vector in \(\mathbb R^m\).

The finite-sample proof in Appendix~C has a gap. The issue is that the
proof uses an empirical-risk-minimization comparison between two
predictors that are produced by different recursive processes. More
precisely, in the notation of \citet{kearns2025networked},
\(\hat y_k^{\mathrm{emp}}\) denotes the predictor produced at node \(k\)
by the empirical recursive process, while \(\hat y_k\) denotes the
predictor produced at node \(k\) by the population recursive process.
The proof writes \(\widehat R(\cdot)\) for the empirical squared loss on
the sampled rows and uses the inequality
\[
    \widehat R(\hat y_k^{\mathrm{emp}})
    \le
    \widehat R(\hat y_k).
\]
The stated justification is that \(\hat y_k^{\mathrm{emp}}\) is the
empirical-risk minimizer at node \(k\). However, this minimization
argument would only apply if \(\hat y_k^{\mathrm{emp}}\) and
\(\hat y_k\) were feasible predictors for the same empirical
least-squares problem.

In the networked recursion, this is not generally the case. At node
\(p_t\), the least-squares problem is formed from the local sample
columns \(X_{S_t}\) together with the prediction vectors received from
the parents of \(p_t\). In the empirical process, these parent inputs are
the previously learned empirical predictions. In the population process,
they are the previously learned population predictions. Therefore the
two recursive processes generally generate different parent inputs, so
the two least-squares problems are built from different linear spans.
Consequently, \(\hat y_k^{\mathrm{emp}}\) is not minimizing the empirical
objective over a class that necessarily contains \(\hat y_k\), and the
displayed inequality does not follow as stated.

\paragraph{Finite-sample analysis.}
Our first contribution is to address the gap in the finite-sample
analysis described above. We give a finite-sample population-risk
guarantee for the full-message protocol without comparing the
sample-based recursive process to a separate population-level recursive
process. The proof works entirely on the realized sample \(D_m\): we
first prove an empirical path theorem comparing the network output to the
empirical all-columns benchmark, that is, the predictor a single learner
would obtain from the same sample if it observed all \(d\) features. We
then apply the uniform generalization bound of
\citet[Theorem~C.2]{kearns2025networked} to the selected network
predictor and to the population all-columns benchmark. This gives the
desired population-risk guarantee while avoiding the invalid
sample-versus-population recursive comparison.

The finite-sample model also makes the communication cost explicit. If
communication is limited to \(k\) numbers per message, a direct approach
is to choose \(k\) public sample rows and run the finite-sample protocol
only on those rows. This reduces communication, but it also reduces the
effective sample size from \(m\) to \(k\). A different approach is to
keep the full \(m\)-sample data set, but send only a \(k\)-dimensional
linear sketch of each \(m\)-dimensional prediction vector.
\paragraph{Shared-sketch protocol.}
Our second contribution is a protocol using a single public sketch matrix
\(S\in\mathbb R^{k\times m}\), drawn independently of the sample.
The difficulty is recursive: each agent fits using its parents' sketched
predictions, so its predictor depends on \(S\) through earlier
regressions. By linearity, every implicit full-sample prediction remains
a combination of the original feature columns, so all residuals lie in
\(V:=\Span\{X_1,\ldots,X_d,y\}\). Conditional on the sample, this
subspace is fixed independently of \(S\) and has dimension at most
\(d+1\). A single oblivious subspace embedding therefore approximately
preserves the empirical squared losses of all linear predictors
simultaneously, including those produced by the recursive protocol.
Thus, for fixed \(d\), embedding distortion, and failure probability,
the required sketch dimension does not grow with network depth.

We also compare the shared-sketch protocol with the row-sampling
baseline. Row sampling uses \(k\) both as the communication budget and as
the statistical sample size, leading to a term of order
\begin{math}
    \widetilde O(d/\sqrt{k}).
\end{math}
Sketching instead uses all \(m\) samples and pays a separate compression
cost. For Gaussian sketches, the leading \(k\)-dependent term is of order
\begin{math}
    O(\Risk(f_A^*)\sqrt{d/k}).
\end{math}
Thus sketching is especially useful when the all-columns benchmark has
small risk. Even in the worst case, our boundedness assumption gives
\begin{math}
    \Risk(f_A^*)\le Y_{\max}^2,
\end{math}
so the sketching term is at most
\begin{math}
    \widetilde O(\sqrt{d/k})
\end{math}
when \(Y_{\max}\) is treated as a constant.

The remainder of the paper is organized as follows.
Section~\ref{sec:preliminaries} introduces the model, the finite-sample
protocol, and the all-columns benchmarks.
Section~\ref{sec:finite-sample} proves the finite-sample guarantee for
the full-message protocol.
Section~\ref{sec:sketching} introduces the shared-sketch protocol,
proves its population-risk guarantee, and compares the resulting bound
with row subsampling.

\input{Related_Work}

%% file: Related_Work.tex
\subsection{Related Work}
\label{subsec:related-work}

Work on information aggregation in networks goes back to the literature on
opinion dynamics and social learning. Classical models such as DeGroot
dynamics and related heuristic update rules study when repeated local
interaction leads to consensus or accurate aggregation of dispersed
information
\citep{degroot1974reaching,balagoyal1998learning,ellisonfudenberg1993rules,golubjackson2010naive,akbarpour2017information}.
Another major line of work studies Bayesian learning and agreement on
networks, including agreement theorems, herding and informational
cascades, and conditions under which local interaction does or does not
aggregate private information efficiently
\citep{aumann1976agreeing,geanakoplos1982cant,aaronson2005complexity,banerjee1992simple,bikhchandani1992theory,acemoglu2011bayesian,mossel2015strategic,mossel2020social,jadbabaie2012nonbayesian,shahrampour2013exponentially,frongillo2023agreement,kong2023false}.
These models typically focus on scalar states, actions, or beliefs,
rather than supervised learning with high-dimensional feature vectors
distributed across agents.

The work most directly related to ours is
\citet{kearns2025networked}, which introduced the DAG-based learning model
that we build on. A closely related recent work,
\citet{bateni2026networkedbinaryclassification}, studies the same
framework for binary classification.
Recent work also establishes optimal population aggregation rates
\citep{bateni2026optimalrates} and studies network design for exact
aggregation \citep{bateni2026optimalnetworks}.
Our paper focuses on a different
direction: the finite-sample and communication-limited aspects of the
squared-loss regression model. In particular, we give a finite-sample
population-risk guarantee for the full-message protocol and introduce a
shared-sketch protocol that compresses the prediction vectors sent across
edges.

Our setting is also related to feature-partitioned collaborative learning,
especially vertical federated learning and split learning. In vertical
federated learning, different parties hold different feature blocks for
the same examples and communicate intermediate quantities in order to
train a joint predictor
\citep{yang2023vflsurvey,khan2022communicationvfl,fu2022celuvfl,castiglia2022compressedvfl,castiglia2023lessvfl,sun2023oneshotvfl,valdeira2024efvfl}.
These works are close in motivation because they also study learning with
distributed features under communication constraints. The difference is
that they focus on federated training protocols, privacy-oriented system
models, or split neural architectures, whereas our model studies
information aggregation through a directed network of learners that
sequentially pass predictions to their descendants.

The sketching part of our paper is related to randomized numerical
linear algebra and sketched least squares. Random projections and
oblivious subspace embeddings have been used extensively to compress
least-squares problems while preserving their geometry
\citep{clarksonwoodruff2013,woodruff2014sketching,pilanciwainwright2015,pilanciwainwright2016ihs}.
Distributed sketching has also been studied as a way to reduce
communication in large-scale optimization
\citep{bartanpilanci2023distributed}. In contrast to these centralized
or standard distributed optimization settings, our sketches are used
inside a recursive networked prediction process.

A related systems-level perspective comes from the literature on
multi-agent and agentic AI workflows. Recent surveys describe agent
workflows as directed graphs or structured computation graphs in which
multiple agents exchange intermediate outputs and coordinate over a
multi-step task
\citep{yu2025agentworkflow,bei2025graphsmeetagents,lin2026workflowoptimization}.
This viewpoint is conceptually related to our model, but that literature
focuses on LLM-based planning, tool use, memory, and workflow design,
whereas our paper studies statistical guarantees for feature-partitioned
supervised learning.

%% file: preliminaries.tex
\section{Preliminaries}
\label{sec:preliminaries}

Let \(\mathcal D\) be a distribution over feature-label pairs \((x,y)\in\mathbb{R}^d\times\mathbb{R}\). For a predictor \(f:\mathbb{R}^d\to\mathbb{R}\), define its population squared loss by
\begin{equation}
    \Risk(f):=\mathbb{E}_{(x,y)\sim\mathcal D}\bigl[(f(x)-y)^2\bigr].
\end{equation}
Let $D_m=\{(x^{(r)},y^{(r)})\}_{r=1}^m$ be \(m\) i.i.d. samples from \(\mathcal D\). We write \(X\in\mathbb{R}^{m\times d}\) for the sample feature matrix and \(y\in\mathbb{R}^m\) for the vector of sampled labels. For a linear predictor \(f_\beta(x)=\beta^\top x\), define
\begin{equation}
    \EmpRisk_m(f_\beta):=\frac1m\norm{X\beta-y}_2^2.
\end{equation}
For a fitted vector \(u\in\mathbb{R}^m\), we also write $\EmpRisk_m(u):=\frac1m\norm{u-y}_2^2$.

\subsection{Network and finite-sample protocol}

Agents are vertices of a directed acyclic graph. Each agent \(i\) observes a subset \(S_i\subseteq[d]\) of the original coordinates. If there is an edge $j \to i$ we call $j$ the \emph{parent} of $i$ and denote the set of all parents of $i$ by \(\Pa(i)\). In the full-message finite-sample protocol, once the parent fitted vectors have been computed, agent \(i\) solves empirical least squares over its own observed sample columns and the fitted vectors received from its parents:
\[
    (\widehat w_i,\widehat a_i)
    \in
    \argmin_{w,a}
    \frac1m\norm{X_{S_i}w+\sum_{j\in\Pa(i)}a_j u_j-y}_2^2.
\]
The fitted vector sent by agent \(i\) on the training sample is $u_i:=X_{S_i}\widehat w_i+\sum_{j\in\Pa(i)}(\widehat a_i)_j u_j$.
Equivalently, at a point \(x\), agent \(i\)'s predictor is the recursively defined linear function
\[
    \widehat f_i(x)=\widehat w_i^\top x_{S_i}+\sum_{j\in\Pa(i)}(\widehat a_i)_j\widehat f_j(x).
\]

We focus on a directed path $P=(p_1,p_2,\ldots,p_D)$, where $p_{t-1}\in\Pa(p_t)\quad\text{for }t=2,\ldots,D$.
For notational simplicity, along this path we write $u_t:=u_{p_t}$, $\widehat f_t:=\widehat f_{p_t}$, and $S_t:=S_{p_t}$.

\subsection{Coverage under random feature allocation}
\label{subsec:random-coverage}

\begin{assumption}[Deterministic \(M\)-coverage]
\label{ass:M-coverage}
We assume every contiguous block of $M$ agents on the path observe the full set of features from $X$. More formally, we have $\bigcup_{t=i}^{i+M-1}S_t=[d]$ for every $i=1,\ldots,D-M+1$.
\end{assumption}

Our proofs use coverage only on the disjoint blocks
\(\{(q-1)M+1,\ldots,qM\}\), \(q=1,\ldots,\lfloor D/M\rfloor\).

The random feature allocation model of
\citet[Section~3.3]{kearns2025networked} gives a sufficient condition for
Assumption~\ref{ass:M-coverage} to hold with high probability.

\begin{corollary}[Coverage under random feature allocation]
\label{cor:random-M-coverage}
Fix a path \(P=(p_1,\ldots,p_D)\) independently of the feature allocation.
Let \(p,\delta_{\mathrm{cover}}\in(0,1)\), and let \(M\) be an integer
with \(1\le M\le D\). Suppose each feature \(\ell\in[d]\) belongs to
\(S_t\) with probability \(p\), independently over all agents on the path
and all features. If
\[
    M\ge
    \frac{\log(dD/\delta_{\mathrm{cover}})}{-\log(1-p)},
\]
then Assumption~\ref{ass:M-coverage} holds with probability at least
\(1-\delta_{\mathrm{cover}}\) over the feature allocation.
\end{corollary}

The proof is given in Appendix~\ref{app:random-coverage}. For fixed
\(p\), the sufficient coverage window grows only logarithmically in
\(d\), \(D\), and \(1/\delta_{\mathrm{cover}}\), provided that this
window fits within the path. Thus the deterministic coverage condition
holds with high probability under this random allocation model.
Coverage may remain restrictive for sparse or dependent feature
allocations. On the coverage event, the results in
Sections~\ref{sec:finite-sample} and~\ref{sec:sketching} apply subject to
their remaining assumptions.

\subsection{All-columns benchmarks}

Fix a radius \(A>0\). The population all-columns benchmark is
\[
    \beta_A^*\in\argmin_{\norm{\beta}_1\le A}\Risk(f_\beta),
    \qquad
    f_A^*(x):=(\beta_A^*)^\top x.
\]
That is, the predictor that a single agent would learn from observing the distribution $\mathcal{D}$. The empirical all-columns benchmark is the predictor that a single agent would learn from the same finite sample if it observed all \(d\) columns:
\[
    \widehat\beta_A\in\argmin_{\norm{\beta}_1\le A}\EmpRisk_m(f_\beta),
    \qquad
    \widehat f_A(x):=\widehat\beta_A^\top x.
\]
The finite-sample proof below first compares the network output to \(\widehat f_A\) on the same sample. Only after this purely empirical comparison do we transfer to the population benchmark \(f_A^*\).

\begin{assumption}[Boundedness and covariance]
\label{ass:bounded}
There are constants \(X_{\max},Y_{\max},\lambda_0>0\) such that
\begin{enumerate}[label=(\roman*),leftmargin=2em]
    \item \(\norm{x}_\infty\le X_{\max}\) almost surely;
    \item \(|y|\le Y_{\max}\) almost surely;
    \item the population covariance matrix \(\Sigma:=\mathbb{E}[xx^\top]\) satisfies \(\lambda_{\min}(\Sigma)\ge\lambda_0\).
\end{enumerate}
\end{assumption}

Boundedness controls the empirical and generalization bounds, while the
covariance condition provides coefficient-norm control for the
population-risk guarantee.
The empirical path argument does not require the covariance condition;
relaxing boundedness or replacing covariance-based norm control would
require new concentration or norm-control arguments.

%% file: finite_sample.tex
\section{Finite-sample analysis}
\label{sec:finite-sample}

This section proves a population-risk guarantee for the full-message
finite-sample protocol from Section~\ref{sec:preliminaries}. We first
state the uniform generalization bound used to pass from empirical risk
to population risk. We then state the empirical path theorem, which is
proved in Appendix~\ref{app:finite-sample-proofs}. Combining these two
results with the norm-control lemma from Appendix~\ref{app:norm-control}
gives the main theorem of this section.

For \(\Lambda>0\), let
\[
    \Hlin_\Lambda
    :=
    \{x\mapsto \beta^\top x:\norm{\beta}_1\le \Lambda\}.
\]

The following theorem follows from the proof of
\citet[Theorem~C.2]{kearns2025networked}.

\begin{theorem}[Generalization bound for \(\ell_1\)-bounded linear regression]
\label{thm:C2}
Fix \(\Lambda>0\). Assume \(\norm{x}_\infty\le X_{\max}\) and
\(|y|\le Y_{\max}\). Let \(D_m\) be a training set of \(m\) i.i.d.\
samples from \(\mathcal D\). Then, for any \(\delta\in(0,1)\), with
probability at least \(1-\delta\) over the draw of \(D_m\), for all
\(f\in\Hlin_\Lambda\),
\[
    \bigl|\Risk(f)-\EmpRisk_m(f)\bigr|
    \le
    \epsCtwo(m,d,\Lambda,\delta),
\]
where
\[
    \epsCtwo(m,d,\Lambda,\delta)
    :=
    C(\Lambda X_{\max}+Y_{\max})\Lambda X_{\max}
    \sqrt{\frac{\log(2d)}{m}}
    +
    (\Lambda X_{\max}+Y_{\max})^2
    \sqrt{\frac{\log(C'/\delta)}{2m}},
\]
and \(C,C'>0\) are universal constants.
\end{theorem}

The quantity \(\epsCtwo(m,d,\Lambda,\delta)\) controls the difference
between empirical risk and population risk uniformly over
\(\Hlin_\Lambda\). We keep both terms in this bound. In particular, after
substituting
\begin{math}
    \Lambda_{\mathrm{net}}=\sqrt{\frac{2d}{\lambda_0}}Y_{\max},
\end{math}
the second term is not lower order in general.

The next theorem is the empirical part of the finite-sample argument. It
compares the network output to the empirical all-columns benchmark
\(\widehat f_A\) on the same realized sample. Its proof is deferred to
Appendix~\ref{app:finite-sample-proofs}.

\begin{theorem}[Empirical path theorem under \(M\)-coverage]
\label{thm:empirical-path}
Let \(K:=\lfloor D/M\rfloor\), and define
\begin{math}
    \Gamma_{\mathrm{path}}
    :=
    2AX_{\max}Y_{\max}\sqrt{\frac{M}{K}}.
\end{math}
Partition the first \(KM\) agents on the path into \(K\) disjoint blocks
of length \(M\). Under Assumption~\ref{ass:M-coverage}, there is a block
endpoint \(b^*\) such that
\[
    \EmpRisk_m(u_{b^*})
    \le
    \EmpRisk_m(\widehat f_A)
    +
    \Gamma_{\mathrm{path}}.
\]
The same bound holds for every later node on the path.
\end{theorem}

Using Theorem~\ref{thm:C2}, Theorem~\ref{thm:empirical-path}, and the
norm-control lemma from Appendix~\ref{app:norm-control}, we obtain the
main result of this section.

\begin{theorem}[Finite-sample theorem for the full-message protocol]
\label{thm:finite-sample-main}
Assume Assumption~\ref{ass:M-coverage} and Assumption~\ref{ass:bounded}.
Let
\begin{math}
    K:=\lfloor D/M\rfloor,
    \Gamma_{\mathrm{path}}:=2AX_{\max}Y_{\max}\sqrt{\frac{M}{K}},
    \Lambda_{\mathrm{net}}:=\sqrt{\frac{2d}{\lambda_0}}Y_{\max},
\end{math}
and let \(\Lambda:=\max\{A,\Lambda_{\mathrm{net}}\}\). On the joint
event that
\begin{enumerate}[label=(\roman*),leftmargin=2em]
    \item \(\lambda_{\min}(m^{-1}X^\top X)\ge\lambda_0/2\), and
    \item the uniform-convergence event of Theorem~\ref{thm:C2} holds for
    \(\Hlin_\Lambda\),
\end{enumerate}
there exists a block endpoint \(b^*\) on the path such that the network
predictor \(\widehat f_{b^*}\) satisfies
\[
    \Risk(\widehat f_{b^*})
    \le
    \Risk(f_A^*)
    +
    \Gamma_{\mathrm{path}}
    +
    2\epsCtwo(m,d,\Lambda,\delta).
\]
The same population-risk bound holds for every later node on the path.
\end{theorem}

The bound separates the aggregation term \(\Gamma_{\mathrm{path}}\)
from the statistical term \(2\epsCtwo(m,d,\Lambda,\delta)\), which
accounts for passing from empirical to population risk. Empirical risk
is nonincreasing along the path, so some block has a small total risk
decrease. Together with the block length, this controls changes in its
fitted values. Coverage and local orthogonality then make the endpoint residual approximately
orthogonal to every feature, yielding the comparison with the
all-columns benchmark. For fixed \(A,X_{\max},Y_{\max},\lambda_0\),
the two terms scale as \(O(M/\sqrt D)\) and
\(\widetilde O(d/\sqrt m)\), respectively. Thus the bound gives a small
excess-risk guarantee when \(D\) is large relative to \(M^2\) and
\(m\) is large enough to make the statistical term small; increasing
depth alone does not eliminate the statistical term.

\begin{proof}
By Theorem~\ref{thm:empirical-path}, there is a block endpoint \(b^*\)
such that
\[
    \EmpRisk_m(u_{b^*})
    \le
    \EmpRisk_m(\widehat f_A)
    +
    \Gamma_{\mathrm{path}}.
\]
By Lemma~\ref{lem:norm-control}, we have
\(\widehat f_{b^*}\in\Hlin_\Lambda\). Also
\(\widehat f_A,f_A^*\in\Hlin_A\subseteq\Hlin_\Lambda\).

Applying Theorem~\ref{thm:C2} to \(\widehat f_{b^*}\),
\[
    \Risk(\widehat f_{b^*})
    \le
    \EmpRisk_m(u_{b^*})
    +
    \epsCtwo(m,d,\Lambda,\delta).
\]
Using the empirical path bound,
\[
    \Risk(\widehat f_{b^*})
    \le
    \EmpRisk_m(\widehat f_A)
    +
    \Gamma_{\mathrm{path}}
    +
    \epsCtwo(m,d,\Lambda,\delta).
\]
Since \(\widehat f_A\) minimizes empirical risk over \(\Hlin_A\) and
\(f_A^*\in\Hlin_A\),
\[
    \EmpRisk_m(\widehat f_A)
    \le
    \EmpRisk_m(f_A^*).
\]
Applying Theorem~\ref{thm:C2} to \(f_A^*\), using
\(\Hlin_A\subseteq\Hlin_\Lambda\), gives
\[
    \EmpRisk_m(f_A^*)
    \le
    \Risk(f_A^*)+\epsCtwo(m,d,\Lambda,\delta).
\]
Combining the last three displays,
\[
    \Risk(\widehat f_{b^*})
    \le
    \Risk(f_A^*)
    +
    \Gamma_{\mathrm{path}}
    +
    2\epsCtwo(m,d,\Lambda,\delta).
\]

For any later node \(t\ge b^*\), Theorem~\ref{thm:empirical-path} gives
the same empirical bound for \(u_t\), and Lemma~\ref{lem:norm-control}
gives \(\widehat f_t\in\Hlin_\Lambda\). Therefore the same argument
applies to \(\widehat f_t\).
\end{proof}

%% file: sketching.tex
\section{Shared-sketch protocol}
\label{sec:sketching}

In this section we introduce the sketching protocol. The training set is
still the same \(m\)-sample data set used in
Section~\ref{sec:finite-sample}. The difference is only in
communication: instead of sending a full \(m\)-dimensional prediction
vector, each agent sends a \(k\)-dimensional sketch of that vector, where
\(k\ll m\).

The main point is that the sample size and the communication size are now
separated. The sample size \(m\) controls the usual statistical
generalization error, while the sketch dimension \(k\) controls how much
information is lost by compression. If the sketch preserves the geometry
of the relevant residual subspace, then the sketched protocol satisfies a
population-risk guarantee similar to the finite-sample theorem, but with
an additional sketching distortion term.

\subsection{Protocol and risks}
\label{subsec:sketch-protocol}

Throughout this section, \(S\in\mathbb R^{k\times m}\) is a public sketch
matrix drawn independently of the sample. For \(v\in\mathbb R^m\), write
\(\widetilde v:=Sv\). Once the sketched parent messages
\(\{\widetilde u_j^{\mathrm{sk}}:j\in\Pa(i)\}\) are available, agent
\(i\) solves the sketched least-squares problem
\[
    (\widetilde w_i,\widetilde a_i)
    \in
    \argmin_{w,a}
    \frac1m
    \norm{
        SX_{S_i}w
        +\sum_{j\in\Pa(i)} a_j\widetilde u_j^{\mathrm{sk}}
        -Sy
    }_2^2.
\]
The agent then broadcasts only the \(k\)-dimensional vector
\[
    \widetilde u_i^{\mathrm{sk}}
    :=
    SX_{S_i}\widetilde w_i
    +\sum_{j\in\Pa(i)}(\widetilde a_i)_j\widetilde u_j^{\mathrm{sk}}
    \in\mathbb R^k.
\]
For the analysis, we also define the corresponding implicit full-sample
fitted vector
\[
    u_i^{\mathrm{sk}}
    :=
    X_{S_i}\widetilde w_i
    +\sum_{j\in\Pa(i)}(\widetilde a_i)_j u_j^{\mathrm{sk}}
    \in\mathbb R^m.
\]
By construction,
\begin{math}
    \widetilde u_i^{\mathrm{sk}}=Su_i^{\mathrm{sk}}.
\end{math}
The vector \(u_i^{\mathrm{sk}}\) is never transmitted. It is only a proof
device that lets us view each \(k\)-dimensional message as the sketch of
a full \(m\)-dimensional prediction vector.

\begin{definition}[Sketched empirical risk]
\label{def:sketched-risk}
For \(u\in\mathbb R^m\), define
\begin{math}
    \SkRisk_{m,S}(u)
    :=
    \frac1m\norm{S(u-y)}_2^2.
\end{math}
For a linear predictor \(f_\beta(x)=\beta^\top x\), define
\begin{math}
    \SkRisk_{m,S}(f_\beta)
    :=
    \frac1m\norm{S(X\beta-y)}_2^2.
\end{math}
If \(\widetilde u=Su\), then
\begin{math}
    \SkRisk_{m,S}(u)
    =
    \frac1m\norm{\widetilde u-Sy}_2^2.
\end{math}
Thus the sketched empirical risk can be computed from the sketched
message \(\widetilde u\) alone.
\end{definition}

Along the fixed path \(P=(p_1,\ldots,p_D)\), we abbreviate
\begin{math}
    u_t^{\mathrm{sk}}:=u_{p_t}^{\mathrm{sk}},
    \widetilde u_t^{\mathrm{sk}}:=Su_t^{\mathrm{sk}},
    S_t:=S_{p_t}.
\end{math}

The all-columns sketched benchmark is
\[
    \widetilde\beta_A
    \in
    \argmin_{\norm{\beta}_1\le A}\SkRisk_{m,S}(f_\beta),
    \qquad
    \widetilde f_A(x):=\widetilde\beta_A^\top x.
\]
This is the predictor that a centralized learner would obtain if it saw
all \(d\) features but minimized the sketched loss instead of the usual
empirical loss.


\subsection{Subspace embeddings}
\label{subsec:ose}

Let
\[
    V:=\Span\{X_1,\ldots,X_d,y\}\subseteq\mathbb R^m,
\]
where \(X_\ell\) is column \(\ell\) of \(X\). Then
\(\dim(V)\le d+1\). This subspace contains every linear prediction
\(X\beta\), and also every residual \(X\beta-y\). In this section the
data matrix \(X\) and label vector \(y\) are fixed, so \(V\) is a fixed
subspace of \(\mathbb R^m\). The randomness in the embedding statements
below is only over the draw of the sketch matrix \(S\).

\begin{definition}[Oblivious subspace embedding]
\label{def:ose}
For \(\varepsilon\in(0,1)\) and a subspace \(W\subseteq\mathbb R^m\), the
matrix \(S\) is an \emph{\((\varepsilon,W)\)-embedding} if for every
\(v\in W\),
\[
    (1-\varepsilon)\norm{v}_2^2
    \le
    \norm{Sv}_2^2
    \le
    (1+\varepsilon)\norm{v}_2^2.
\]
\end{definition}

An \((\varepsilon,W)\)-embedding approximately preserves the Euclidean
length of every vector in \(W\). Since our risks are squared residual
norms, embedding the residual subspace is exactly what is needed to
compare empirical risk and sketched empirical risk.

\begin{lemma}[Sketch preserves residual norms]
\label{lem:sketch-preserves-residuals}
Let \(\mathcal V\subseteq\mathbb R^m\), and suppose \(S\) is an
\((\varepsilon,W)\)-embedding for some
\(W\supseteq\Span(\mathcal V\cup\{y\})\). Then for every
\(u\in\Span(\mathcal V)\),
\[
    (1-\varepsilon)\EmpRisk_m(u)
    \le
    \SkRisk_{m,S}(u)
    \le
    (1+\varepsilon)\EmpRisk_m(u).
\]
\end{lemma}

\begin{proof}
Since \(u\in\Span(\mathcal V)\), the residual \(u-y\) lies in \(W\).
Apply Definition~\ref{def:ose} to \(u-y\) and divide by \(m\).
\end{proof}

The implicit predictors produced by the sketched protocol remain linear
functions of the original feature columns, by the same induction used for
Lemma~\ref{lem:empirical-linearity}. Hence all vectors
\(u_t^{\mathrm{sk}}\) lie in \(\Span\{X_1,\ldots,X_d\}\subseteq V\).
Therefore an \((\varepsilon,V)\)-embedding is enough for the whole
sketching analysis.

A standard way to obtain such an embedding is to use a Gaussian sketch:
each entry of \(S\in\mathbb R^{k\times m}\) is drawn independently from
\(N(0,1/k)\). The next lemma is the only random-sketch fact we need. It
is a direct consequence of the standard Gaussian oblivious-subspace
embedding theorem; see, for example,
\citet[Section~2.1]{woodruff2014sketching} for subspace embeddings and
\citet[Theorem~5.39]{vershynin2012nonasymptotic} for the Gaussian
singular-value estimate behind the proof. The general subspace embedding
statement and proof are deferred to Appendix~\ref{app:sketching-proofs}.

\begin{lemma}[Gaussian embedding for the residual subspace]
\label{lem:gaussian-ose-V}
Fix \(X\in\mathbb R^{m\times d}\) and \(y\in\mathbb R^m\), and let
\[
    V=\Span\{X_1,\ldots,X_d,y\}.
\]
Let \(S\in\mathbb R^{k\times m}\) have i.i.d.\ entries \(N(0,1/k)\).
There is a universal constant \(C_{\mathrm{ose}}>0\) such that, if
\[
    k
    \ge
    C_{\mathrm{ose}}\,
    \frac{d+1+\log(1/\delta_{\mathrm{ose}})}{\varepsilon^2},
\]
then \(S\) is an \((\varepsilon,V)\)-embedding with probability at least
\(1-\delta_{\mathrm{ose}}\) over the draw of \(S\).
\end{lemma}

In the deterministic theorem below, we simply assume that \(S\) is an
\((\varepsilon,V)\)-embedding. Lemma~\ref{lem:gaussian-ose-V} shows that
this event holds with high probability for Gaussian sketches provided
\[
    k\gtrsim
    \frac{d+\log(1/\delta_{\mathrm{ose}})}{\varepsilon^2}.
\]

\subsection{Main guarantees for the shared-sketch protocol}
\label{subsec:sketch-guarantees}

The proof follows the same idea as the finite-sample proof. We need two
facts: sketched risk should be close enough to population risk, and along
the path there should be a node whose sketched risk is close to the
sketched all-columns benchmark.

The first theorem converts sketched empirical risk into population risk,
and also converts population risk back into sketched empirical risk. Its
proof is deferred to Appendix~\ref{app:sketching-proofs}.

\begin{theorem}[Sketched generalization bound]
\label{thm:sketched-C2}
Fix \(\Lambda>0\) and assume Assumption~\ref{ass:bounded}(i)--(ii).
Suppose the uniform-convergence event of Theorem~\ref{thm:C2} holds for
\(\Hlin_\Lambda\), and that \(S\) is an \((\varepsilon,V)\)-embedding
with \(\varepsilon\in(0,1)\). Then, simultaneously for all
\(f\in\Hlin_\Lambda\),
\begin{align}
\label{eq:sketched-gen-bound-upper}
    \Risk(f)
    &\le
    \frac{1}{1-\varepsilon}\,\SkRisk_{m,S}(f)
    +\epsCtwo(m,d,\Lambda,\delta), \\
\label{eq:sketched-gen-bound-lower}
    \SkRisk_{m,S}(f)
    &\le
    (1+\varepsilon)\,\Risk(f)
    +(1+\varepsilon)\,\epsCtwo(m,d,\Lambda,\delta).
\end{align}
\end{theorem}

The second theorem is the sketched analogue of the empirical path
theorem. It says that some block endpoint has sketched risk close to the
best all-columns sketched predictor. Its proof is also deferred to
Appendix~\ref{app:sketching-proofs}.

\begin{theorem}[Sketched empirical path theorem]
\label{thm:sketched-path}
Assume Assumption~\ref{ass:M-coverage}. Let \(K:=\lfloor D/M\rfloor\)
and partition the first \(KM\) nodes of the path into \(K\) disjoint
blocks of length \(M\). Let
\begin{math}
    \Gamma_{\mathrm{path}}
    :=
    2AX_{\max}Y_{\max}\sqrt{\frac{M}{K}}.
\end{math}
On the event that \(S\) is an \((\varepsilon,V)\)-embedding, there is a
block endpoint \(b^*\) such that
\[
    \SkRisk_{m,S}(u_{b^*}^{\mathrm{sk}})
    \le
    \SkRisk_{m,S}(\widetilde f_A)
    +(1+\varepsilon)\Gamma_{\mathrm{path}}.
\]
The same bound holds with \(\widetilde f_A\) replaced by \(f_A^*\).
\end{theorem}

Combining Theorem~\ref{thm:sketched-C2} and
Theorem~\ref{thm:sketched-path} gives the main theorem for the
shared-sketch protocol.

\begin{theorem}[Finite-sample theorem for the shared-sketch protocol]
\label{thm:sketched-main}
Assume Assumption~\ref{ass:M-coverage} and Assumption~\ref{ass:bounded}.
Let
\begin{math}
    K:=\lfloor D/M\rfloor,
    \Gamma_{\mathrm{path}}:=2AX_{\max}Y_{\max}\sqrt{\frac{M}{K}}.
\end{math}
Define
\begin{math}
    \Lambda_{\mathrm{sk}}(\varepsilon)
    :=
    \sqrt{\frac{2d}{\lambda_0}}
    \left(
        1+\sqrt{\frac{1+\varepsilon}{1-\varepsilon}}
    \right)Y_{\max},
\end{math}
and let
\begin{math}
    \Lambda:=\max\{A,\Lambda_{\mathrm{sk}}(\varepsilon)\}.
\end{math}
On the joint event that
\begin{enumerate}[label=(\roman*),leftmargin=2em]
    \item \(S\) is an \((\varepsilon,V)\)-embedding,
    \item \(\lambda_{\min}(m^{-1}X^\top X)\ge\lambda_0/2\), and
    \item the uniform-convergence event of Theorem~\ref{thm:C2} holds
    for \(\Hlin_\Lambda\),
\end{enumerate}
there exists a block endpoint \(b^*\) such that the implicit linear
predictor \(\widetilde f_{b^*}(x):=\widetilde\beta_{b^*}^\top x\)
satisfies
\[
    \Risk(\widetilde f_{b^*})
    \le
    \frac{1+\varepsilon}{1-\varepsilon}\,\Risk(f_A^*)
    +\frac{(1+\varepsilon)\,\Gamma_{\mathrm{path}}}{1-\varepsilon}
    +\frac{2}{1-\varepsilon}\,\epsCtwo(m,d,\Lambda,\delta).
\]
The same bound holds for every later node \(t\ge b^*\) on the path.
\end{theorem}

\begin{remark}[Multiplicative versus additive form]
\label{rem:multiplicative-bound}
The leading sketching correction in Theorem~\ref{thm:sketched-main}
scales with \(\Risk(f_A^*)\), the risk of the benchmark predictor. If we
instead used a purely additive bound, the sketching error would scale
with the worst-case quantity
\((\Lambda X_{\max}+Y_{\max})^2\). The multiplicative form is therefore
at least as good in the worst case and can be much better when
\(f_A^*\) has small risk.
\end{remark}

\begin{proof}[Proof of Theorem~\ref{thm:sketched-main}]
By Theorem~\ref{thm:sketched-path}, there is a block endpoint \(b^*\)
such that
\[
    \SkRisk_{m,S}(u_{b^*}^{\mathrm{sk}})
    \le
    \SkRisk_{m,S}(f_A^*)
    +(1+\varepsilon)\Gamma_{\mathrm{path}}.
\]
By Lemma~\ref{lem:sketched-norm-control},
\[
    u_{b^*}^{\mathrm{sk}}=X\widetilde\beta_{b^*},
    \qquad
    \norm{\widetilde\beta_{b^*}}_1
    \le
    \Lambda_{\mathrm{sk}}(\varepsilon)
    \le
    \Lambda.
\]
Hence \(\widetilde f_{b^*}\in\Hlin_\Lambda\). Also
\(f_A^*\in\Hlin_A\subseteq\Hlin_\Lambda\), so
Theorem~\ref{thm:sketched-C2} applies to both predictors.

Since
\[
    \SkRisk_{m,S}(\widetilde f_{b^*})
    =
    \SkRisk_{m,S}(u_{b^*}^{\mathrm{sk}}),
\]
Theorem~\ref{thm:sketched-C2} gives
\[
    \Risk(\widetilde f_{b^*})
    \le
    \frac{1}{1-\varepsilon}
    \SkRisk_{m,S}(u_{b^*}^{\mathrm{sk}})
    +
    \epsCtwo(m,d,\Lambda,\delta).
\]
Using the sketched path bound,
\[
    \Risk(\widetilde f_{b^*})
    \le
    \frac{1}{1-\varepsilon}
    \Bigl(
        \SkRisk_{m,S}(f_A^*)
        +(1+\varepsilon)\Gamma_{\mathrm{path}}
    \Bigr)
    +
    \epsCtwo(m,d,\Lambda,\delta).
\]
Applying Theorem~\ref{thm:sketched-C2} again to \(f_A^*\),
\[
    \SkRisk_{m,S}(f_A^*)
    \le
    (1+\varepsilon)\Risk(f_A^*)
    +(1+\varepsilon)\epsCtwo(m,d,\Lambda,\delta).
\]
Substituting,
\[
    \Risk(\widetilde f_{b^*})
    \le
    \frac{1+\varepsilon}{1-\varepsilon}\Risk(f_A^*)
    +
    \frac{(1+\varepsilon)\Gamma_{\mathrm{path}}}{1-\varepsilon}
    +
    \frac{2}{1-\varepsilon}\epsCtwo(m,d,\Lambda,\delta).
\]

For any later node \(t\ge b^*\), Lemma~\ref{lem:sketched-norm-control}
gives \(\widetilde f_t\in\Hlin_\Lambda\), and sketched monotonicity gives
\[
    \SkRisk_{m,S}(u_t^{\mathrm{sk}})
    \le
    \SkRisk_{m,S}(u_{b^*}^{\mathrm{sk}}).
\]
The same argument therefore gives the same population-risk bound for
\(\widetilde f_t\).
\end{proof}
\subsection{Comparison with row subsampling}
\label{subsec:sketched-comparison}

We briefly compare the shared-sketch protocol with row subsampling at the
same per-edge communication budget \(k\). In row subsampling, agents keep
\(k\) public rows and run the finite-sample protocol on those rows.
Thus the communicated vector has length \(k\), but the effective sample
size in the generalization term is also \(k\). On the analogous
covariance and uniform-convergence events for the \(k\)-row sample,
Theorem~\ref{thm:finite-sample-main} gives
\[
    \Risk(\widehat f_{b^*})-\Risk(f_A^*)
    \le
    O(\Gamma_{\mathrm{path}})
    +
    \widetilde O\!\left(\frac{d}{\sqrt{k}}\right),
\]
where \(X_{\max},Y_{\max}\), and \(\lambda_0\) are treated as constants.

For Gaussian sketches, Lemma~\ref{lem:gaussian-ose-V} and
Theorem~\ref{thm:sketched-main} give, for
\[
    \varepsilon
    \asymp
    \sqrt{\frac{d+\log(1/\delta_{\mathrm{ose}})}{k}},
\]
the bound
\[
    \Risk(\widetilde f_{b^*})-\Risk(f_A^*)
    \le
    O(\Gamma_{\mathrm{path}})
    +
    \widetilde O\!\left(\frac{d}{\sqrt m}\right)
    +
    O\!\left(
        \Risk(f_A^*)
        \sqrt{\frac{d+\log(1/\delta_{\mathrm{ose}})}{k}}
    \right),
\]
up to lower-order terms. The key difference is that row sampling pays a
statistical price at sample size \(k\), while sketching keeps the
statistical term at sample size \(m\) and pays a separate compression
cost. Subject to the embedding requirement, a useful budget makes the
compression term no larger than \(\Gamma_{\mathrm{path}}\), provided
the statistical term is already small.

In many natural regimes, the all-columns benchmark risk
\(\Risk(f_A^*)\) is small. For example, this happens when the best
all-features linear predictor is nearly realizable, or when the noise
around the best all-features predictor is low. In such regimes, the
leading \(k\)-dependent term in the sketched bound,
\begin{math}
     O\!\left(
        \Risk(f_A^*)
        \sqrt{\frac{d+\log(1/\delta_{\mathrm{ose}})}{k}}
    \right),
\end{math}
is much smaller than the row-sampling term
\begin{math}
    \widetilde O\!\left(\frac{d}{\sqrt{k}}\right).
\end{math}
Thus, in the typical regime where the all-columns benchmark is accurate,
the sketch bound is expected to be better than the row-sampling bound.
When \(\Risk(f_A^*)\) is large, matching this benchmark need not imply
small absolute prediction error.

Even in the worst case, since \(0\in\Hlin_A\),
\begin{math}
    \Risk(f_A^*)\le \Risk(0)=\mathbb E[y^2]\le Y_{\max}^2.
\end{math}
Therefore, when \(Y_{\max}\) is treated as a constant, the leading
\(k\)-dependent term in the sketched bound is at most
\begin{math}
     \widetilde O\!\left(\sqrt{\frac{d}{k}}\right),
\end{math}
whereas row sampling gives
\begin{math}
    \widetilde O\!\left(\frac{d}{\sqrt{k}}\right).
\end{math}
So, at the level of the stated upper bounds, sketching has better leading
dependence on \(d\) and \(k\), while also separating the finite-sample
term \(\widetilde O(d/\sqrt m)\) from the communication-dependent term.
See Appendix~\ref{app:sketching-comparison} for a more detailed
comparison.

The data-generating relationship need not be linear; the benchmark and
local learners are linear. Extending the protocol to adaptive nonlinear
learners remains future work, since their residuals need not lie in one
fixed low-dimensional subspace.

%% file: appendix.tex
\appendix

\section{Deferred proofs for the finite-sample analysis}
\label{app:finite-sample-proofs}

This appendix proves Theorem~\ref{thm:empirical-path} and
Lemma~\ref{lem:norm-control}. These results are deterministic statements
about the realized sample \(D_m\). The final subsection proves
Corollary~\ref{cor:random-M-coverage}, which concerns random feature
allocation.

\subsection{Least-squares identities}

We first collect the basic identities used in the empirical path proof.

\begin{lemma}[Linearity of sequential empirical predictors]
\label{lem:empirical-linearity}
For every agent \(p_t\) on the path, there exists
\(\widehat\beta_t\in\mathbb R^d\) such that
\[
    u_t=X\widehat\beta_t.
\]
Equivalently, the implicit predictor
\(\widehat f_t(x)=\widehat\beta_t^\top x\) is a linear function of the
original \(d\) features.
\end{lemma}

\begin{proof}
Proceed by induction over the topological order of the graph. At a
source node, \(u_i\) is a linear combination of the columns of
\(X_{S_i}\). At a non-source node,
\[
    u_i
    =
    X_{S_i}\widehat w_i
    +
    \sum_{j\in\Pa(i)}(\widehat a_i)_j u_j.
\]
By induction, each parent vector \(u_j\) is a linear combination of the
columns of \(X\). Therefore \(u_i\) is also a linear combination of the
columns of \(X\).
\end{proof}

For each path node \(p_t\), define
\[
    Z_t:=\bigl[\,X_{S_t}\quad u_j:j\in\Pa(p_t)\,\bigr],
    \qquad
    r_t:=y-u_t.
\]

\begin{lemma}[Empirical orthogonality]
\label{lem:empirical-orthogonality}
For every \(t\),
\[
    Z_t^\top r_t=0.
\]
Consequently,
\begin{enumerate}[label=(\roman*),leftmargin=2em]
    \item for every coordinate \(\ell\in S_t\), \(X_\ell^\top r_t=0\);
    \item for \(t\ge2\), \(u_{t-1}^\top r_t=0\);
    \item \(u_t^\top r_t=0\).
\end{enumerate}
\end{lemma}

\begin{proof}
By definition, \(u_t=Z_t\widehat\theta_t\), where
\(\widehat\theta_t\) minimizes
\[
    \frac1m\norm{Z_t\theta-y}_2^2.
\]
The normal equations give
\[
    Z_t^\top(Z_t\widehat\theta_t-y)=0,
\]
which is equivalent to \(Z_t^\top r_t=0\). The three consequences follow
because \(X_\ell\), \(u_{t-1}\), and \(u_t\) all lie in \(\Span(Z_t)\)
whenever the corresponding statement applies.
\end{proof}

\begin{lemma}[Empirical monotonicity along the path]
\label{lem:empirical-monotonicity}
For every \(t\ge2\),
\[
    \EmpRisk_m(u_t)\le \EmpRisk_m(u_{t-1}).
\]
\end{lemma}

\begin{proof}
Agent \(p_t\) can copy \(p_{t-1}\) by placing coefficient \(1\) on
\(u_{t-1}\) and coefficient \(0\) on all other parent predictions and
local features. Since \(u_t\) is chosen by least squares, its empirical
risk cannot be larger.
\end{proof}

\begin{lemma}[Empirical Pythagorean identity]
\label{lem:empirical-pythagorean}
For every \(t\ge2\),
\[
    \frac1m\norm{u_t-u_{t-1}}_2^2
    =
    \EmpRisk_m(u_{t-1})-\EmpRisk_m(u_t).
\]
\end{lemma}

\begin{proof}
Since \(y-u_{t-1}=r_t+(u_t-u_{t-1})\),
\[
    \norm{y-u_{t-1}}_2^2
    =
    \norm{r_t}_2^2
    +
    \norm{u_t-u_{t-1}}_2^2
    +
    2\ip{r_t}{u_t-u_{t-1}}.
\]
Both \(u_t\) and \(u_{t-1}\) lie in \(\Span(Z_t)\), so
Lemma~\ref{lem:empirical-orthogonality} gives
\(\ip{r_t}{u_t-u_{t-1}}=0\). Dividing by \(m\) proves the claim.
\end{proof}

\begin{lemma}[Empirical squared-loss decomposition]
\label{lem:loss-decomposition}
For any \(u,g\in\mathbb R^m\), with \(r:=y-u\),
\[
    \EmpRisk_m(u)
    =
    \EmpRisk_m(g)
    +
    \frac{2}{m}\ip{g}{r}
    -
    \frac{2}{m}\ip{u}{r}
    -
    \frac1m\norm{u-g}_2^2.
\]
\end{lemma}

\begin{proof}
Since \(y-g=(y-u)+(u-g)=r+(u-g)\),
\[
    \norm{y-g}_2^2
    =
    \norm{r}_2^2
    +
    2\ip{r}{u-g}
    +
    \norm{u-g}_2^2.
\]
Dividing by \(m\) and rearranging gives
\[
    \EmpRisk_m(u)
    =
    \EmpRisk_m(g)
    -
    \frac2m\ip{r}{u-g}
    -
    \frac1m\norm{u-g}_2^2.
\]
Expanding \(\ip{r}{u-g}=\ip{u}{r}-\ip{g}{r}\) proves the identity.
\end{proof}

\subsection{The empirical block theorem and proof of the empirical path theorem}

For a contiguous block \(B=[a,b]\) of nodes on the path, define
\[
    \mathcal U_B
    :=
    \{X_\ell:\ell\in\textstyle\bigcup_{t=a}^b S_t\}
    \cup
    \{u_j:j\in\Pa(p_t),\ t\in[a,b]\}.
\]
For a representation \(g=\sum_{v\in\mathcal U_B}\alpha_v v\), define
\[
    \norm{\alpha}_{1,B}
    :=
    \sum_{v\in\mathcal U_B}
    |\alpha_v|\,\frac{\norm{v}_2}{\sqrt m}.
\]

\begin{theorem}[Empirical block theorem]
\label{thm:empirical-block}
Let \(B=[a,b]\subseteq\{1,\ldots,D\}\) be a contiguous block on the path,
and let
\[
    \Delta_B:=\EmpRisk_m(u_a)-\EmpRisk_m(u_b)\ge0.
\]
Let \(g=\sum_{v\in\mathcal U_B}\alpha_v v\). Then
\[
    \EmpRisk_m(u_b)
    \le
    \EmpRisk_m(g)
    +
    2\norm{\alpha}_{1,B}\sqrt{|B|\Delta_B}.
\]
\end{theorem}

\begin{proof}
Fix \(v\in\mathcal U_B\). By definition, there is
\(t=t(v)\in[a,b]\) such that \(v\) is either a local feature column
available at \(p_t\) or a parent prediction available at \(p_t\). Hence
\(v\in\Span(Z_t)\), and Lemma~\ref{lem:empirical-orthogonality} gives
\[
    v^\top(y-u_t)=0.
\]
Therefore
\[
    \frac1m v^\top(y-u_b)
    =
    \frac1m v^\top(u_t-u_b).
\]
By Cauchy--Schwarz,
\[
    \left|
    \frac1m v^\top(y-u_b)
    \right|
    \le
    \frac{\norm{v}_2}{\sqrt m}
    \cdot
    \frac{\norm{u_b-u_t}_2}{\sqrt m}.
\]
Using
\[
    u_b-u_t=\sum_{s=t+1}^b (u_s-u_{s-1}),
\]
the triangle inequality, Lemma~\ref{lem:empirical-pythagorean}, and
Cauchy--Schwarz give
\[
\begin{aligned}
    \frac{\norm{u_b-u_t}_2}{\sqrt m}
    &\le
    \sum_{s=t+1}^b
    \frac{\norm{u_s-u_{s-1}}_2}{\sqrt m} \\
    &=
    \sum_{s=t+1}^b
    \sqrt{\EmpRisk_m(u_{s-1})-\EmpRisk_m(u_s)} \\
    &\le
    \sqrt{
        (b-t)
        \sum_{s=a+1}^b
        \bigl(\EmpRisk_m(u_{s-1})-\EmpRisk_m(u_s)\bigr)
    } \\
    &\le
    \sqrt{|B|\Delta_B}.
\end{aligned}
\]
Thus
\[
    \left|
    \frac1m v^\top(y-u_b)
    \right|
    \le
    \frac{\norm{v}_2}{\sqrt m}\sqrt{|B|\Delta_B}.
\]
Multiplying by \(|\alpha_v|\) and summing over \(v\in\mathcal U_B\),
\[
    \left|
    \frac1m\ip{g}{y-u_b}
    \right|
    \le
    \norm{\alpha}_{1,B}\sqrt{|B|\Delta_B}.
\]
Now apply Lemma~\ref{lem:loss-decomposition} with \(u=u_b\). By
Lemma~\ref{lem:empirical-orthogonality}, \(u_b^\top(y-u_b)=0\), and the
term \(-m^{-1}\norm{u_b-g}_2^2\) is nonpositive. Therefore
\[
    \EmpRisk_m(u_b)
    \le
    \EmpRisk_m(g)
    +
    2\norm{\alpha}_{1,B}\sqrt{|B|\Delta_B}.
\]
\end{proof}

\begin{proof}[Proof of Theorem~\ref{thm:empirical-path}]
By Lemma~\ref{lem:empirical-monotonicity}, the empirical risks are
nonincreasing. Hence the total drop across the \(K\) disjoint blocks is
at most
\[
    \EmpRisk_m(u_1)\le \EmpRisk_m(0)\le Y_{\max}^2.
\]
Therefore one block \(B^*=[a^*,b^*]\) satisfies
\[
    \Delta_{B^*}\le \frac{Y_{\max}^2}{K}.
\]
By Assumption~\ref{ass:M-coverage}, the features in \(B^*\) cover all of
\([d]\). Thus
\[
    \widehat f_A=X\widehat\beta_A,
    \qquad
    \norm{\widehat\beta_A}_1\le A,
\]
is representable in \(\mathcal U_{B^*}\) using only raw feature columns.
For each raw feature column \(X_\ell\),
\[
    \frac{\norm{X_\ell}_2}{\sqrt m}\le X_{\max}.
\]
Hence the representation of \(\widehat f_A\) satisfies
\[
    \norm{\widehat\beta_A}_{1,B^*}\le AX_{\max}.
\]
Applying Theorem~\ref{thm:empirical-block} with
\(g=\widehat f_A\) and \(|B^*|=M\), we get
\[
    \EmpRisk_m(u_{b^*})
    \le
    \EmpRisk_m(\widehat f_A)
    +
    2AX_{\max}Y_{\max}\sqrt{\frac{M}{K}}.
\]
The bound for later nodes follows from
Lemma~\ref{lem:empirical-monotonicity}.
\end{proof}

\subsection{Norm control}
\label{app:norm-control}

To apply Theorem~\ref{thm:C2} to the network predictors, we need one
common \(\ell_1\)-ball that contains all predictors produced along the
path. The next lemma gives such a bound under the empirical covariance
event.

\begin{lemma}[\(\ell_1\)-norm control]
\label{lem:norm-control}
Assume
\[
    \lambda_{\min}(m^{-1}X^\top X)\ge\lambda_0/2.
\]
Then every network predictor on the path admits a representation
\(u_t=X\widehat\beta_t\) with
\[
    \norm{\widehat\beta_t}_2
    \le
    \sqrt{\frac{2}{\lambda_0}}\,Y_{\max},
    \qquad
    \norm{\widehat\beta_t}_1
    \le
    \Lambda_{\mathrm{net}}
    :=
    \sqrt{\frac{2d}{\lambda_0}}\,Y_{\max}.
\]
\end{lemma}

\begin{proof}
By Lemma~\ref{lem:empirical-linearity}, write
\(u_t=X\widehat\beta_t\). Since \(u_t\) is the orthogonal projection of
\(y\) onto \(\Span(Z_t)\),
\[
    \norm{u_t}_2\le\norm{y}_2.
\]
Therefore
\[
    \widehat\beta_t^\top
    \left(\frac1mX^\top X\right)
    \widehat\beta_t
    =
    \frac1m\norm{u_t}_2^2
    \le
    \frac1m\norm{y}_2^2
    \le
    Y_{\max}^2.
\]
The empirical covariance lower bound gives
\[
    \norm{\widehat\beta_t}_2
    \le
    \sqrt{\frac{2}{\lambda_0}}\,Y_{\max}.
\]
The \(\ell_1\)-bound follows from
\[
    \norm{\widehat\beta_t}_1
    \le
    \sqrt d\,\norm{\widehat\beta_t}_2.
\]
\end{proof}

\subsection{Proof of the random-coverage corollary}
\label{app:random-coverage}

\begin{proof}[Proof of Corollary~\ref{cor:random-M-coverage}]
Fix a feature \(\ell\in[d]\) and a block of \(M\) consecutive agents on
\(P\). By independence across agents, the probability that none of these
agents observes \(\ell\) is \((1-p)^M\). There are \(D-M+1\) such
blocks and \(d\) features. A union bound over all feature--block pairs
gives
\[
\begin{aligned}
    \Pr\!\left[\text{Assumption~\ref{ass:M-coverage} fails on }P\right]
    &\le d(D-M+1)(1-p)^M \\
    &\le dD(1-p)^M \\
    &\le \delta_{\mathrm{cover}},
\end{aligned}
\]
where the last inequality follows from the stated lower bound on \(M\).
The union bound does not require independence between overlapping
blocks.
\end{proof}

\section{Deferred proofs for the shared-sketch protocol}
\label{app:sketching-proofs}

This appendix contains the deferred proofs for
Section~\ref{sec:sketching}.

\subsection{Gaussian subspace embeddings}

We first record the standard Gaussian oblivious-subspace-embedding
statement used in Lemma~\ref{lem:gaussian-ose-V}.

\begin{lemma}[Gaussian sketches give subspace embeddings]
\label{lem:gaussian-ose}
Let \(W\subseteq\mathbb R^m\) be a fixed subspace with
\(\dim(W)=r\). Let \(S\in\mathbb R^{k\times m}\) have i.i.d.\ entries
\(N(0,1/k)\). There is a universal constant \(C_{\mathrm{ose}}>0\) such
that, if
\[
    k
    \ge
    C_{\mathrm{ose}}\,
    \frac{r+\log(1/\delta_{\mathrm{ose}})}{\varepsilon^2},
\]
then \(S\) is an \((\varepsilon,W)\)-embedding with probability at least
\(1-\delta_{\mathrm{ose}}\) over the draw of \(S\).
\end{lemma}

\begin{proof}
Let \(Q\in\mathbb R^{m\times r}\) have orthonormal columns spanning
\(W\). Every \(v\in W\) can be written as \(v=Qz\) for some
\(z\in\mathbb R^r\), and \(\norm{v}_2=\norm{z}_2\). Therefore \(S\) is
an \((\varepsilon,W)\)-embedding if and only if
\[
    (1-\varepsilon)\norm{z}_2^2
    \le
    \norm{SQz}_2^2
    \le
    (1+\varepsilon)\norm{z}_2^2
    \qquad
    \text{for all }z\in\mathbb R^r.
\]
Equivalently,
\[
    \norm{Q^\top S^\top S Q-I_r}_{\mathrm{op}}\le\varepsilon.
\]
Since \(Q\) is fixed and has orthonormal columns, \(SQ\) is a
\(k\times r\) Gaussian matrix with i.i.d.\ entries \(N(0,1/k)\). The
standard non-asymptotic singular-value bound for Gaussian matrices gives
\[
    \norm{Q^\top S^\top S Q-I_r}_{\mathrm{op}}\le\varepsilon
\]
with probability at least \(1-\delta_{\mathrm{ose}}\), provided
\[
    k
    \ge
    C_{\mathrm{ose}}\,
    \frac{r+\log(1/\delta_{\mathrm{ose}})}{\varepsilon^2}.
\]
This proves the claim.
\end{proof}

\begin{proof}[Proof of Lemma~\ref{lem:gaussian-ose-V}]
The subspace
\[
    V=\Span\{X_1,\ldots,X_d,y\}
\]
is fixed and has dimension at most \(d+1\). Apply
Lemma~\ref{lem:gaussian-ose} with \(W=V\) and \(r=d+1\).
\end{proof}

\subsection{Proof of the sketched generalization theorem}

\begin{proof}[Proof of Theorem~\ref{thm:sketched-C2}]
Fix \(f_\beta\in\Hlin_\Lambda\). Since
\begin{math}
    V=\Span\{X_1,\ldots,X_d,y\},
\end{math}
we have \(X\beta-y\in V\). Hence, by
Lemma~\ref{lem:sketch-preserves-residuals},
\begin{equation}
\label{eq:sketched-C2-embedding-step}
    (1-\varepsilon)\EmpRisk_m(f_\beta)
    \le
    \SkRisk_{m,S}(f_\beta)
    \le
    (1+\varepsilon)\EmpRisk_m(f_\beta).
\end{equation}
On the uniform-convergence event of Theorem~\ref{thm:C2},
\begin{equation}
\label{eq:sketched-C2-uniform-step}
    \bigl|\Risk(f_\beta)-\EmpRisk_m(f_\beta)\bigr|
    \le
    \epsCtwo(m,d,\Lambda,\delta).
\end{equation}

By \eqref{eq:sketched-C2-uniform-step},
\begin{math}
    \Risk(f_\beta)
    \le
    \EmpRisk_m(f_\beta)+\epsCtwo(m,d,\Lambda,\delta).
\end{math}
By the left inequality in \eqref{eq:sketched-C2-embedding-step},
\begin{math}
    \EmpRisk_m(f_\beta)
    \le
    \frac{1}{1-\varepsilon}\SkRisk_{m,S}(f_\beta).
\end{math}
Therefore
\begin{math}
    \Risk(f_\beta)
    \le
    \frac{1}{1-\varepsilon}\SkRisk_{m,S}(f_\beta)
    +\epsCtwo(m,d,\Lambda,\delta),
\end{math}
which proves \eqref{eq:sketched-gen-bound-upper}.

Similarly, by the right inequality in
\eqref{eq:sketched-C2-embedding-step},
\begin{math}
    \SkRisk_{m,S}(f_\beta)
    \le
    (1+\varepsilon)\EmpRisk_m(f_\beta).
\end{math}
Using \eqref{eq:sketched-C2-uniform-step} again,
\begin{math}
    \EmpRisk_m(f_\beta)
    \le
    \Risk(f_\beta)+\epsCtwo(m,d,\Lambda,\delta).
\end{math}
Thus
\begin{math}
    \SkRisk_{m,S}(f_\beta)
    \le
    (1+\varepsilon)\Risk(f_\beta)
    +(1+\varepsilon)\epsCtwo(m,d,\Lambda,\delta),
\end{math}
which proves \eqref{eq:sketched-gen-bound-lower}. Since both events hold
uniformly over \(f_\beta\in\Hlin_\Lambda\), the two inequalities hold
simultaneously for all \(f\in\Hlin_\Lambda\).
\end{proof}

\subsection{Proof of the sketched empirical path theorem}

The sketched protocol is the empirical least-squares protocol applied to
the compressed data matrix \(SX\) and compressed label vector \(Sy\).
Therefore the proof of Theorem~\ref{thm:empirical-block} applies
verbatim after replacing every vector \(v\) by its sketch \(Sv\), and
after measuring dictionary size by the normalized sketched norm
\(\norm{Sv}_2/\sqrt m\).

For a block \(B=[a,b]\), define
\[
    \Delta_B^{\mathrm{sk}}
    :=
    \SkRisk_{m,S}(u_a^{\mathrm{sk}})
    -
    \SkRisk_{m,S}(u_b^{\mathrm{sk}}).
\]
Let
\[
    \mathcal U_B^{\mathrm{sk}}
    :=
    \{X_\ell:\ell\in\textstyle\bigcup_{t=a}^b S_t\}
    \cup
    \{u_j^{\mathrm{sk}}:j\in\Pa(p_t),\ t\in[a,b]\}.
\]
If \(g=\sum_{v\in\mathcal U_B^{\mathrm{sk}}}\alpha_v v\), define
\[
    \norm{\alpha}_{1,S,B}
    :=
    \sum_{v\in\mathcal U_B^{\mathrm{sk}}}
    |\alpha_v|\,\frac{\norm{Sv}_2}{\sqrt m}.
\]
The sketched version of Theorem~\ref{thm:empirical-block} is
\begin{equation}
\label{eq:sketched-block-bound}
    \SkRisk_{m,S}(u_b^{\mathrm{sk}})
    \le
    \SkRisk_{m,S}(g)
    +
    2\norm{\alpha}_{1,S,B}\sqrt{|B|\Delta_B^{\mathrm{sk}}}.
\end{equation}
This is the same block theorem applied in sketch space.

\begin{proof}[Proof of Theorem~\ref{thm:sketched-path}]
Partition the first \(KM\) nodes into \(K\) disjoint blocks of length
\(M\). The sketched risk is nonincreasing along the path, since each node
can copy the sketched prediction of its path predecessor. Hence
\[
    \sum_{\text{blocks }B}\Delta_B^{\mathrm{sk}}
    \le
    \SkRisk_{m,S}(u_1^{\mathrm{sk}})
    \le
    \SkRisk_{m,S}(0)
    =
    \frac1m\norm{Sy}_2^2.
\]
Since \(y\in V\) and \(S\) is an \((\varepsilon,V)\)-embedding,
\[
    \frac1m\norm{Sy}_2^2
    \le
    (1+\varepsilon)\frac1m\norm{y}_2^2
    \le
    (1+\varepsilon)Y_{\max}^2.
\]
Thus there exists a block \(B^*=[a^*,b^*]\) such that
\[
    \Delta_{B^*}^{\mathrm{sk}}
    \le
    \frac{(1+\varepsilon)Y_{\max}^2}{K}.
\]

By Assumption~\ref{ass:M-coverage}, the features in \(B^*\) cover all of
\([d]\). Therefore
\[
    \widetilde f_A=X\widetilde\beta_A,
    \qquad
    \norm{\widetilde\beta_A}_1\le A,
\]
is representable in \(\mathcal U_{B^*}^{\mathrm{sk}}\) using only raw
feature columns. For every feature column \(X_\ell\), the embedding event
gives
\[
    \frac{\norm{SX_\ell}_2}{\sqrt m}
    \le
    \sqrt{1+\varepsilon}\,
    \frac{\norm{X_\ell}_2}{\sqrt m}
    \le
    \sqrt{1+\varepsilon}\,X_{\max}.
\]
Hence
\[
    \norm{\widetilde\beta_A}_{1,S,B^*}
    \le
    \sqrt{1+\varepsilon}\,AX_{\max}.
\]
Applying \eqref{eq:sketched-block-bound} with \(g=\widetilde f_A\) and
\(|B^*|=M\), we get
\[
\begin{aligned}
    \SkRisk_{m,S}(u_{b^*}^{\mathrm{sk}})
    &\le
    \SkRisk_{m,S}(\widetilde f_A)
    +
    2\sqrt{1+\varepsilon}\,AX_{\max}
    \sqrt{
        M\cdot
        \frac{(1+\varepsilon)Y_{\max}^2}{K}
    } \\
    &=
    \SkRisk_{m,S}(\widetilde f_A)
    +(1+\varepsilon)\Gamma_{\mathrm{path}}.
\end{aligned}
\]
Finally, since \(\widetilde f_A\) minimizes \(\SkRisk_{m,S}\) over
\(\Hlin_A\) and \(f_A^*\in\Hlin_A\),
\[
    \SkRisk_{m,S}(\widetilde f_A)
    \le
    \SkRisk_{m,S}(f_A^*).
\]
Thus the same bound also holds with \(\widetilde f_A\) replaced by
\(f_A^*\).
\end{proof}

\subsection{Linearity, monotonicity, and norm control}
\label{app:sketched-norm-control}

To apply Theorem~\ref{thm:sketched-C2} to the output of the sketched
protocol, we need to know that the implicit sketched predictors are
linear predictors with bounded \(\ell_1\)-norm. We also use the
monotonicity of sketched risk along the path.

\begin{lemma}[Linearity, monotonicity, and norm control for sketched predictors]
\label{lem:sketched-norm-control}
On the joint event that \(S\) is an \((\varepsilon,V)\)-embedding and
\[
    \lambda_{\min}(m^{-1}X^\top X)\ge \lambda_0/2,
\]
every implicit sketched predictor on the path admits a representation
\[
    u_t^{\mathrm{sk}}=X\widetilde\beta_t
\]
with
\[
    \norm{\widetilde\beta_t}_1
    \le
    \Lambda_{\mathrm{sk}}(\varepsilon)
    :=
    \sqrt{\frac{2d}{\lambda_0}}
    \left(
        1+\sqrt{\frac{1+\varepsilon}{1-\varepsilon}}
    \right)Y_{\max}.
\]
Moreover, the sketched empirical risk is nonincreasing along the path:
\[
    \SkRisk_{m,S}(u_t^{\mathrm{sk}})
    \le
    \SkRisk_{m,S}(u_{t-1}^{\mathrm{sk}})
    \qquad
    \text{for every } t\ge2.
\]
\end{lemma}

\begin{proof}
First we prove linearity. The proof is the same induction as in
Lemma~\ref{lem:empirical-linearity}. At a source node, the implicit
sketched predictor is a linear combination of columns of \(X_{S_i}\). At
a non-source node,
\[
    u_i^{\mathrm{sk}}
    =
    X_{S_i}\widetilde w_i
    +
    \sum_{j\in\Pa(i)}(\widetilde a_i)_j u_j^{\mathrm{sk}}.
\]
By induction, each parent predictor \(u_j^{\mathrm{sk}}\) is a linear
combination of columns of \(X\). Therefore \(u_i^{\mathrm{sk}}\) is also
a linear combination of columns of \(X\). Hence for every path node \(t\),
there exists \(\widetilde\beta_t\in\mathbb R^d\) such that
\[
    u_t^{\mathrm{sk}}=X\widetilde\beta_t.
\]

Next we prove sketched monotonicity. Since \(p_{t-1}\in\Pa(p_t)\), agent
\(p_t\) can copy the sketched prediction of its path predecessor by
placing coefficient \(1\) on \(Su_{t-1}^{\mathrm{sk}}\) and coefficient
\(0\) on all other inputs. Since \(p_t\) minimizes the sketched
least-squares objective, this feasible choice implies
\[
    \SkRisk_{m,S}(u_t^{\mathrm{sk}})
    \le
    \SkRisk_{m,S}(u_{t-1}^{\mathrm{sk}}).
\]

It remains to prove the coefficient bound. Since
\(u_t^{\mathrm{sk}}\in\Span\{X_1,\ldots,X_d\}\subseteq V\), the residual
\(y-u_t^{\mathrm{sk}}\) lies in \(V\). By
Lemma~\ref{lem:sketch-preserves-residuals},
\[
    (1-\varepsilon)\EmpRisk_m(u_t^{\mathrm{sk}})
    \le
    \SkRisk_{m,S}(u_t^{\mathrm{sk}}).
\]
By sketched monotonicity and the fact that the zero predictor is feasible
at the first node,
\[
    \SkRisk_{m,S}(u_t^{\mathrm{sk}})
    \le
    \SkRisk_{m,S}(0)
    =
    \frac1m\norm{Sy}_2^2.
\]
Since \(y\in V\) and \(S\) is an \((\varepsilon,V)\)-embedding,
\[
    \frac1m\norm{Sy}_2^2
    \le
    (1+\varepsilon)\frac1m\norm{y}_2^2
    \le
    (1+\varepsilon)Y_{\max}^2.
\]
Therefore
\[
    \frac1{\sqrt m}\norm{y-u_t^{\mathrm{sk}}}_2
    \le
    \sqrt{\frac{1+\varepsilon}{1-\varepsilon}}\,Y_{\max}.
\]
By the triangle inequality,
\[
    \frac1{\sqrt m}\norm{u_t^{\mathrm{sk}}}_2
    \le
    \left(
        1+\sqrt{\frac{1+\varepsilon}{1-\varepsilon}}
    \right)Y_{\max}.
\]
Since \(u_t^{\mathrm{sk}}=X\widetilde\beta_t\),
\[
    \widetilde\beta_t^\top
    \left(\frac1mX^\top X\right)
    \widetilde\beta_t
    =
    \frac1m\norm{u_t^{\mathrm{sk}}}_2^2.
\]
Using
\[
    \lambda_{\min}(m^{-1}X^\top X)\ge\lambda_0/2,
\]
we get
\[
    \norm{\widetilde\beta_t}_2
    \le
    \sqrt{\frac{2}{\lambda_0}}
    \left(
        1+\sqrt{\frac{1+\varepsilon}{1-\varepsilon}}
    \right)Y_{\max}.
\]
Finally,
\[
    \norm{\widetilde\beta_t}_1
    \le
    \sqrt d\,\norm{\widetilde\beta_t}_2
    \le
    \sqrt{\frac{2d}{\lambda_0}}
    \left(
        1+\sqrt{\frac{1+\varepsilon}{1-\varepsilon}}
    \right)Y_{\max}
    =
    \Lambda_{\mathrm{sk}}(\varepsilon).
\]
\end{proof}

\subsection{Detailed comparison with row subsampling}
\label{app:sketching-comparison}

We give the detailed comparison between row subsampling and Gaussian
sketching. The main text states the leading-order conclusion; here we
spell out the bounds.

First consider the row-sampling baseline. In this protocol, agents choose
\(k\) public rows and run the full-message finite-sample protocol on only
those rows. Thus the message length is \(k\), but the sample size in the
uniform-convergence term is also \(k\). On the analogous empirical
covariance and uniform-convergence events for the \(k\)-row sample,
Theorem~\ref{thm:finite-sample-main} gives
\[
    \Risk(\widehat f_{b^*})-\Risk(f_A^*)
    \le
    \Gamma_{\mathrm{path}}
    +
    2\epsCtwo(k,d,\bar\Lambda_{\mathrm{net}},\delta),
\]
where
\[
    \bar\Lambda_{\mathrm{net}}
    :=
    \max\left\{A,\sqrt{\frac{2d}{\lambda_0}}Y_{\max}\right\}.
\]
By definition,
\[
    \epsCtwo(k,d,\bar\Lambda_{\mathrm{net}},\delta)
    =
    C(\bar\Lambda_{\mathrm{net}}X_{\max}+Y_{\max})\bar\Lambda_{\mathrm{net}}X_{\max}
    \sqrt{\frac{\log(2d)}{k}}
    +
    (\bar\Lambda_{\mathrm{net}}X_{\max}+Y_{\max})^2
    \sqrt{\frac{\log(C'/\delta)}{2k}}.
\]
If \(A,X_{\max},Y_{\max}\), and \(\lambda_0\) are treated as fixed
constants, then
\[
    \bar\Lambda_{\mathrm{net}}=\Theta(\sqrt d).
\]
Hence the first term is
\[
    \widetilde O\!\left(\frac d{\sqrt k}\right),
\]
and the second term is
\[
    \widetilde O\!\left(\frac d{\sqrt k}\right).
\]
Thus the row-sampling guarantee has the
leading-order form
\[
    \Risk(\widehat f_{b^*})-\Risk(f_A^*)
    \le
    O(\Gamma_{\mathrm{path}})
    +
    \widetilde O\!\left(\frac d{\sqrt k}\right).
\]

Now consider Gaussian sketches. By Lemma~\ref{lem:gaussian-ose-V}, if
\[
    k
    \gtrsim
    \frac{d+\log(1/\delta_{\mathrm{ose}})}{\varepsilon^2},
\]
then \(S\) is an \((\varepsilon,V)\)-embedding with probability at least
\(1-\delta_{\mathrm{ose}}\). For \(\varepsilon\le 1/2\),
\[
    \sqrt{\frac{1+\varepsilon}{1-\varepsilon}}\le \sqrt 3.
\]
Therefore
\[
    \Lambda_{\mathrm{sk}}(\varepsilon)
    \le
    \sqrt{\frac{2d}{\lambda_0}}(1+\sqrt 3)Y_{\max}.
\]
It is enough to use the fixed radius
\[
    \bar\Lambda_{\mathrm{sk}}
    :=
    \max\left\{
        A,\,
        \sqrt{\frac{2d}{\lambda_0}}(1+\sqrt 3)Y_{\max}
    \right\}.
\]

Theorem~\ref{thm:sketched-main} gives
\[
    \Risk(\widetilde f_{b^*})
    \le
    \frac{1+\varepsilon}{1-\varepsilon}\Risk(f_A^*)
    +
    \frac{(1+\varepsilon)\Gamma_{\mathrm{path}}}{1-\varepsilon}
    +
    \frac{2}{1-\varepsilon}
    \epsCtwo(m,d,\bar\Lambda_{\mathrm{sk}},\delta).
\]
Subtracting \(\Risk(f_A^*)\) and using
\[
    \frac{1+\varepsilon}{1-\varepsilon}=1+O(\varepsilon),
    \qquad
    \frac{1}{1-\varepsilon}=O(1)
    \qquad
    \text{for }\varepsilon\le 1/2,
\]
we get
\[
\begin{aligned}
    \Risk(\widetilde f_{b^*})-\Risk(f_A^*)
    \le\;&
    O(\Gamma_{\mathrm{path}})
    +
    O\!\left(\epsCtwo(m,d,\bar\Lambda_{\mathrm{sk}},\delta)\right) \\
    &+
    O\!\left(
        \varepsilon
        \bigl[
            \Risk(f_A^*)
            +\Gamma_{\mathrm{path}}
            +\epsCtwo(m,d,\bar\Lambda_{\mathrm{sk}},\delta)
        \bigr]
    \right).
\end{aligned}
\]
Taking
\[
    \varepsilon
    \asymp
    \sqrt{\frac{d+\log(1/\delta_{\mathrm{ose}})}{k}},
\]
this becomes
\[
\begin{aligned}
    \Risk(\widetilde f_{b^*})-\Risk(f_A^*)
    \le\;&
    O(\Gamma_{\mathrm{path}})
    +
    O\!\left(\epsCtwo(m,d,\bar\Lambda_{\mathrm{sk}},\delta)\right) \\
    &+
    O\!\left(
        \bigl[
            \Risk(f_A^*)
            +\Gamma_{\mathrm{path}}
            +\epsCtwo(m,d,\bar\Lambda_{\mathrm{sk}},\delta)
        \bigr]
        \sqrt{\frac{d+\log(1/\delta_{\mathrm{ose}})}{k}}
    \right).
\end{aligned}
\]

Since \(\bar\Lambda_{\mathrm{sk}}=\Theta(\sqrt d)\) under the same
constant-parameter convention,
\[
    \epsCtwo(m,d,\bar\Lambda_{\mathrm{sk}},\delta)
    =
    \widetilde O\!\left(\frac d{\sqrt m}\right).
\]
Thus the sketched guarantee can be summarized as
\[
    \Risk(\widetilde f_{b^*})-\Risk(f_A^*)
    \le
    O(\Gamma_{\mathrm{path}})
    +
    \widetilde O\!\left(\frac d{\sqrt m}\right)
    +
    O\!\left(
        \Risk(f_A^*)
        \sqrt{\frac{d+\log(1/\delta_{\mathrm{ose}})}{k}}
    \right),
\]
up to lower-order terms in which
\(\Gamma_{\mathrm{path}}\) and
\(\epsCtwo(m,d,\bar\Lambda_{\mathrm{sk}},\delta)\) are also multiplied by
the sketching distortion.

This highlights the difference between the two protocols. Row sampling
uses \(k\) both as the communication budget and as the statistical sample
size, leading to the term
\[
    \widetilde O\!\left(\frac d{\sqrt k}\right).
\]
Sketching uses all \(m\) samples and pays a separate compression cost.
The leading \(k\)-dependent term is scaled by the benchmark risk
\(\Risk(f_A^*)\). In many natural regimes this risk is small, for example
when the all-features linear benchmark is nearly realizable or when the
noise level around the best all-features predictor is low.

Even in the worst case, the benchmark risk is bounded by the risk of the
zero predictor. Since \(0\in\Hlin_A\),
\[
    \Risk(f_A^*)
    \le
    \Risk(0)
    =
    \mathbb E[y^2]
    \le
    Y_{\max}^2.
\]
Therefore, when \(Y_{\max}\) is treated as a constant, the leading
\(k\)-dependent term in the sketched bound is at most
\[
    \widetilde O\!\left(\sqrt{\frac d k}\right).
\]
Thus, at the level of leading-order dimension and communication
dependence, sketching replaces the row-sampling term
\[
    \widetilde O\!\left(\frac d{\sqrt k}\right)
\]
by
\[
    \widetilde O\!\left(\sqrt{\frac d k}\right),
\]
while keeping the finite-sample term separate as
\[
    \widetilde O\!\left(\frac d{\sqrt m}\right).
\]
This comparison is at the level of upper bounds; constants, logarithmic
factors, and lower-order terms can still matter for finite values of
\(d,m,k\).